\documentclass[11pt,a4paper]{article}
\usepackage[hyperref]{naaclhlt2019}
\usepackage{times}
\usepackage{latexsym}
\usepackage{color}
\usepackage{url}
\usepackage{graphicx}
\usepackage{here}

\usepackage{lingmacros}
\usepackage{bm}
\usepackage{caption}
\usepackage{subcaption}
\usepackage{booktabs}
\usepackage{amsfonts}
\usepackage{CJKutf8}
\usepackage{amsmath, amsfonts}
\usepackage{xcolor}
\usepackage{comment}
\usepackage{threeparttable}
\usepackage{cleveref}

\usepackage{amsmath,amssymb,mathtools}
\usepackage{tikz}
\usetikzlibrary{positioning}
\usepackage[framemethod=TikZ]{mdframed}
\mdfsetup{%
 middlelinewidth=0,
 backgroundcolor=gray!15,
 roundcorner=8pt
}

\aclfinalcopy 

\title{Cross-Corpora Evaluation and Analysis of Grammatical Error Correction Models --- Is Single-Corpus Evaluation Enough?}

\author{Masato Mita$^{1,2}$,\hspace{-0.2em} Tomoya Mizumoto$^{1}$,\hspace{-0.15em} Masahiro Kaneko$^{3,1}$,\hspace{-0.15em} Ryo Nagata$^{4,1}$,\hspace{-0.15em} Kentaro Inui$^{2,1}$ \hspace{-0.2em}\\
 $^{1}$RIKEN AIP, $^{2}$Tohoku University, $^{3}$Tokyo Metropolitan University, $^{4}$Konan University \\
 {\tt \{masato.mita, tomoya.mizumoto\}@riken.jp,} \hspace{-0.5em} \\
 {\tt kaneko-masahiro@ed.tmu.ac.jp,} \hspace{-0.5em} 
 {\tt nagata-naacl@ml.hyogo-u.ac.jp., }\hspace{-0.5em} \\
 {\tt inui@ecei.tohoku.ac.jp}\hspace{-0.5em}
 }
 
\date{}

\begin{document}
\maketitle
\begin{abstract}
This study explores the necessity of performing cross-corpora evaluation for grammatical error correction (GEC) models. 
GEC models have been previously evaluated based on a single commonly applied corpus: the CoNLL-2014 benchmark.
However, the evaluation remains incomplete because the task difficulty varies depending on the test corpus and conditions such as the proficiency levels of the writers and essay topics. 
To overcome this limitation, we evaluate the performance of several GEC models, including NMT-based (LSTM, CNN, and transformer) and an SMT-based model, against various learner corpora (CoNLL-2013, CoNLL-2014, FCE, JFLEG, ICNALE, and KJ).
Evaluation results reveal that the models' rankings considerably vary depending on the corpus, indicating that single-corpus evaluation is insufficient for GEC models.

\end{abstract}

\section{Introduction}
\label{sec:Intro}
Grammatical error correction (GEC) is the task of correcting various grammatical errors in a given text, which is typically written by non-native speakers. 
Previous studies focused on typical errors such as those in the use of articles~\cite{Han:06:Journal}, prepositions~\cite{Felice:08:COLING}, and noun numbers \cite{Nagata:06:COLING}.
Machine translation approaches are being presently applied for GEC \cite{Junczys-Dowmunt:18:NAACL,Chollampatt:18:AAAI,Ge:18:arXiv,Junczys-Dowmunt:16:EMNLP}.
In these approaches, GEC is treated as a translation problem from the erroneous text to the correct text~\cite{Mizumoto:12:COLING,Felice:14:CoNLLST,Junczysdowmunt:14:CoNLLST}.

However, the evaluation of GEC performance is unfortunately not complete because researchers tend to evaluate their models on a single corpus.
The CoNLL-2014 shared task dataset~\cite{Ng:14:CoNLLST} has been recently used for such evaluation.

Single-corpus evaluation may be insufficient in cases wherein a GEC model generally aims to robustly correct grammatical errors in {\em any\/} written text partly because the task difficulty varies depending on proficiency levels and essay topics.
Although a model outperforms a baseline in one corpus, the model in another corpus may perform better, leading to different conclusions from what we know.
This study explores the necessity of performing cross-corpora evaluation for GEC models.
The performance of four recent models, namely three neural machine translation (NMT)-based models (LSTM, CNN, and transformer) and a statistical machine translation (SMT)-based model is evaluated against six learner corpora (CoNLL-2014, CoNLL-2013~\cite{Ng:13:CoNLLST}, FCE~\cite{Yannakoudakis:11:ACL}, JFLEG~\cite{Napoles:17:EACL}, KJ~\cite{Nagata:11:ACL}, and ICNLAE~\cite{Ishikawa:13:Journal}).
Evaluation results show that the models' rankings considerably vary depending on the corpus.
Empirical results reveal that models must be evaluated using multiple corpora from different perspectives.

The contributions of this study are as follows:

\begin{itemize}
\setlength{\parskip}{0.05cm}
\setlength{\itemsep}{0.05cm}
\item We first explore the necessity of performing cross-corpora evaluation for GEC models.
\item We empirically show that the single-corpus evaluation may be unreliable.
\item Our source code is published for cross-corpora evaluation so that researchers in the community can adequately and easily evaluate their models based on multiple corpora.~\footnote{~\url{https://github.com/tomo-wb/GEC_CCE}}
\end{itemize}

\section{Related Work}
We are motivated by the issue of robustness in the parsing community. 
This field previously focused on improving parsing accuracy on Penn Treebank~\cite{Marcus:93:Journal}.
However, robustness was largely improved by evaluation using multiple corpora including Ontonotes~\cite{Hovy:06:NAACL} and Google Web Treebank~\cite{Petrov:12:SANCL}.
A situation similar to this might also occur in GEC.
In other words, evaluation in GEC has relied heavily on the CoNLL-2014 benchmark, which implies that the field is overdeveloping on this dataset.

Other corpora are used for evaluation, such as KJ \cite{Mizumoto:12:COLING} and JFLEG \cite{Sakaguch:17:IJCNLP,Junczys-Dowmunt:18:NAACL,Chollampatt:18:AAAI,Ge:18:arXiv,Xie:18:NAACL}.
However, these corpora still depend on one or at most two corpora.

\section{Experimental Setup}

\begin{table*}[t]
 \centering
 \scalebox{0.93}[0.93]{ 
 \begin{tabular}{lrcrcccc}
 \toprule
 Corpus & \# sent. & \# refs. & WER & \# topics & Multiple L1 & Multiple proficiency & Public available \\ \midrule
 CoNLL-2014 & 1,312 & 2 & 12.35 & 2 & No & No & Yes \\
 CoNLL-2013 & 1,381 & 1 & 14.85 & 2 & No & No & Yes \\
 FCE & 32,199 & 1 & 12.00 & 10 & Yes & Yes & Yes \\
 JFLEG & 747 & 4 & 20.86 & Many & Yes & Yes & Yes \\
 KJ & 3,081 & 1 & 13.53 & 10 & No & No & Yes \\
 ICNALE & 1,736 & 1 & 7.64 & 2 & Yes & Yes & No \\
 \bottomrule
 \end{tabular}
}
 \caption{Properties of evaluation corpora. Yes/No indicates whether the corpus exhibits each property in terms of multiple L1, multiple proficiency and public available.}
 \label{statics}
\end{table*}


\subsection{Corpora for Evaluation}
\label{subsec:datasets}
Cross-corpora evaluation is discussed herein using six corpora, namely CoNLL-2014, CoNLL-2013, FCE, JFLEG, KJ, and ICNALE.
The following conditions were considered when selecting corpora:

\begin{itemize}
\item The corpus must be used at least once in the GEC community.
\item Based on the hypothesis that writers' proficiency affects the error distribution of any given text, we add a corpus with relatively low proficiency (KJ) compared to CoNLL-2014.
\end{itemize}

We explicitly describe each learner corpus as follows:

\noindent\paragraph{\bf CoNLL-2014~\cite{Ng:14:CoNLLST}}, the official dataset of CoNLL-2014 shared task, is a collection of essays written by students at the National University of Singapore and is commonly used as test data for the CoNLL-2014 benchmark. This dataset contains only two essay topics.

\noindent\paragraph{\bf CoNLL-2013~\cite{Ng:13:CoNLLST}}, the official dataset of CoNLL-2013 shared tasks, is commonly used as the development data for the CoNLL-2014 benchmark and contains only two essay topics.

\noindent\paragraph{\bf Cambridge ESOL First Certificate in English (FCE)~\cite{Yannakoudakis:11:ACL}} is a dataset containing 1,244 examination scripts of the Cambridge FCE examination. Topics and first languages (L1s) in the dataset are diversified because it contains essays for 10 topics written by non-native speakers from various countries.

\noindent\paragraph{\bf JHU FLuency-Extended GUG Corpus (JFLEG)~\cite{Napoles:17:EACL}} contains approximately 1,500 sentences from an English proficiency test. It contains sentences written by learners of the English language having various L1s and proficiency levels.

\noindent\paragraph{\bf Konan-JIEM Learner Corpus (KJ)~\cite{Nagata:11:ACL}} contains 233 essays written on 10 topics by students of a Japanese college, which are manually error-tagged and shallow-parsed.

\noindent\paragraph{\bf International Corpus Network of Asian Learners of English, Written Essays (ICNALE)~\cite{Ishikawa:13:Journal}} contains essays written by college and graduate students from ten Asian countries/regions (China, Hong Kong, Indonesia, Japan, Korea, Pakistan, the Philippines, Singapore, Taiwan, and Thailand). The original ICNALE is not error annotated. Therefore, we sampled a total number of 1,736 sentences, which are manually annotated with grammatical errors based on KJ 's annotation scheme.

Table \ref{statics} summarizes the properties of these corpora.
Let $N$ and $M$ denote the total number of source words and sentences in a corpus, respectively. 
Word error rate (WER) is defined as follows:

\[
\mbox{WER} = \frac{\sum_{m=1}^{M} d(X^{m},Y^{m})}{\sum_{m=1}^{M} N^{m}}
\] \vspace{2mm}

where $X^{m}$ denotes each source sentence, $Y^{m}$ denotes each corrected sentence, and $d(X^{m}, Y^{m})$ denotes the edit distance between $X^{m}$ and $Y^{m}$ using dynamic programming.

The following conclusions are derived: (1) CoNLL-2014 has narrow coverage of topics, proficiency and L1s compared with other corporas such as JFLEG and FCE. (2) Several learner corpora are available for the evaluation of GEC models. These corpora can help investigate the performance of GEC models under different conditions.

\subsection{Models}

The following factors are considered while selecting our model.

\begin{itemize}
\item The models must be recent and commonly used.
\item Each model must be implemented to have a competitive performance on CoNLL-2014.
\end{itemize}

We employed the following models based on the aforementioned factors:

\noindent\paragraph{\bf LSTM}: 
We use a bi-directional LSTM in the encoder and an LSTM with an attention mechanism in the decoder. Both the encoder and the decoder comprise two layers. The LSTM hidden state and word embedding sizes are set to be 500.

\noindent\paragraph{\bf CNN}: 
We follow the previous study \cite{Chollampatt:18:AAAI}, namely a fully convolutional encoder-decoder architecture with seven convolutional layers.
The hyperparameters used in a previous study are used herein \cite{Chollampatt:18:AAAI}.

\noindent\paragraph{\bf Transformer}: 
Transformer is the self-attention-based model proposed by \newcite{Vaswani:17:NIPS}. 
Six layers are used for both the encoder and decoder along with eight attention heads. 
The word embedding size is set to 1024 dimensions, and the size of position-wise feed-forward networks is set to 4096 dimensions at each inner layer.

\noindent\paragraph{\bf SMT}: 
We essentially follow the idea used in a previous study \cite{Junczys-Dowmunt:16:EMNLP}, with some key differences. 
Specifically, we only use English Wikipedia for language model training and only the NUS Corpus of Learner
English (NUCLE) and the Lang-8 Learner Corpora (Lang-8) for translation model training to make the experimental settings equal in all models.

\subsection{Experimental Settings}

We use two public datasets, namely Lang-8 \cite{Mizumoto:11:IJCNLP} and NUCLE \cite{Dahlmeier:13:BEA}, for training. 
Our pre-processing and experimental setup is similar to that reported previously \cite{Chollampatt:18:AAAI}. 
In particular, a subset of NUCLE (5.4K) is utilized as the development data for selecting the model; the remaining subset (1.3M) is utilized as the training data. 
All the models are trained, tuned, and tested in the same way.
The models are tested on each test data shown in Table \ref{statics}. 
As an evaluation metric, we use $F_{0.5}$ score computed by applying the MaxMatch scorer \cite{Dahlmeier:12:NAACL} and GLEU \cite{Napoles:15:ACL}. 
We determine the average $F_{0.5}$ and average GLEU scores of the four models, which are trained with different random initializations, following a previously reported approach \cite{Chollampatt:18:AAAI}.

\begin{figure*}[t]
 \centering\includegraphics[width=1.0\linewidth]{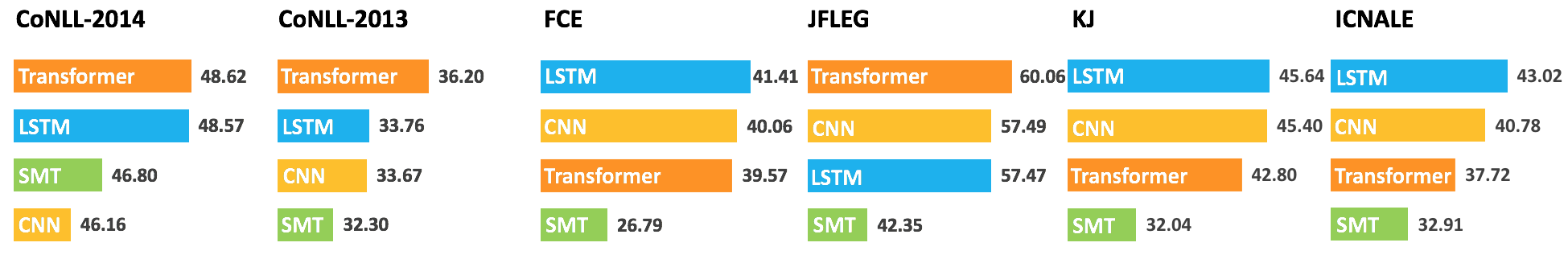}
 \caption{ Average $F_{0.5}$ of the four models (trained with different random initializations), ranked best to worst. }
 \label{f05_ranking}
\end{figure*}
\begin{figure*}[t]
 \centering\includegraphics[width=1.0\linewidth]{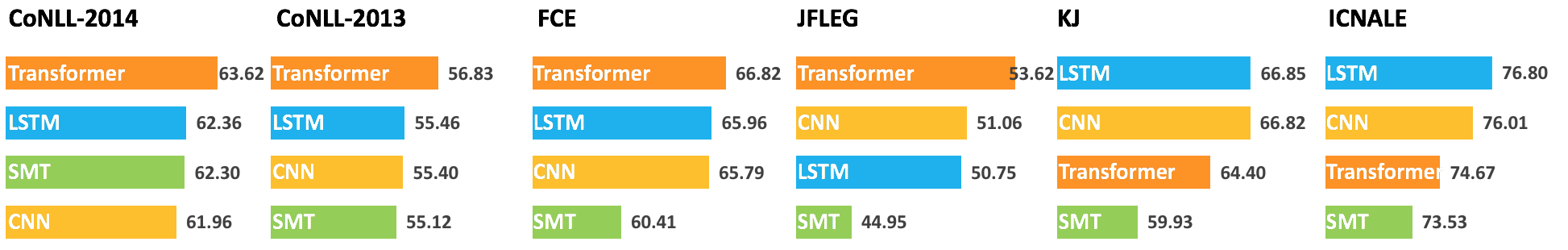}
 \caption{Average GLEU of the four models (trained with different random initializations), ranked best to worst. }
 \label{gleu_ranking}
\end{figure*}


\section{Cross-Corpora Evaluation}
\label{sec:result}

Figure~\ref{f05_ranking} shows the performance of each model sorted from best to worst based on their $F_{0.5}$ score, revealing that the performance substantially varies depending on the corpus. 
For example, the performance of the transformer ranges from the score of $F_{0.5}$, which is as low as 36.20 on CoNLL-2013, to as high as 60.06 on JFLEG. 
Notably, their rankings also considerably vary.
Transformer performs best on CoNLL-2014. 
However, it exhibits third-best performance among FCE, KJ, and ICNALE; LSTM outperforms the other models by a large margin of up to 5.3 $F_{0.5}$ points. 
Some examples of the model outputs are presented in Table \ref{example_Conll-2014} and Table \ref{example_KJ}. 
Some situations are successfully corrected using transformer (Table \ref{example_Conll-2014}), whereas it failed to perform in other situations (Table \ref{example_KJ}). 
The reason for difference in the model rankings cannot be generally stated because it is influenced by various factors such as the learner's proficiency, essay topic, and L1. 
The experimental results show, however, that discussions based on the performance on CoNLL-2014 may only hold under certain conditions.

Figure \ref{gleu_ranking} shows the performance measured in GLEU having a similar trend. However, their rankings on FCE show different trends in Figure \ref{f05_ranking} and Figure \ref{gleu_ranking}.
This is partly because $F_{0.5}$ and GLEU evaluate different perspectives of the models. Furthermore, evaluation data and metric must be appropriately set depending on the factors that need to be evaluated in the model.

\section{Discussion}
\label{sec:analysis}


\begin{table*}[t]
\centering
\begin{tabular}{|l|p{13cm}|}
\hline
 & Sentence \\ \hline
Source & Hence , some \textit{seen} it as being considerate in keeping the genetic risk of getting the disease \textit{in} confidential . \\ \hline
Reference & Hence , some \textbf{see} it as being considerate in keeping the genetic risk of getting the disease \textbf{[DEL]} confidential . \\ \hline \hline
1. Transformer & Hence , some \textbf{see} it as being considerate in keeping the genetic risk of getting the disease \textbf{[DEL]} confidential . \\ \hline
2. LSTM & Hence , some seen it as being considerate in keeping the genetic risk of getting the disease in \textbf{confidentiality} . \\ \hline
3. CNN & Hence , some seen it as being considerate in keeping the genetic risk of getting the disease in \textbf{confidentiality} . \\ \hline
\end{tabular}
\caption{Examples of model outputs on CoNLL-2014.} 
\label{example_Conll-2014}
\end{table*}


\begin{table*}[t]
\centering
\begin{tabular}{|l|p{13cm}|}
\hline
 & Sentence \\ \hline
Source & \textit{In} that day , the time I left school was about eleven p.m . \\ \hline
Reference & \textbf{On} that day , the time I left school was about eleven p.m . \\ \hline \hline
1. LSTM & \textbf{On} that day , the time I left school was about eleven p.m . \\ \hline
2. CNN & \textbf{On} that day , the time I left school was about eleven p.m . \\ \hline
3. Transformer & \textbf{That} day , the time I left school was about eleven p.m .\\ \hline
\end{tabular}
\caption{Examples of model outputs on KJ.} 
\label{example_KJ}
\end{table*}

\begin{table}[t]
 \centering
 \scalebox{0.75}[0.75]{ 
\begin{tabular}{l|ccc|ccc}
\toprule
WER (\%) & \multicolumn{3}{c|}{Low (7.64)} & \multicolumn{3}{c}{High (20.86)} \\ \midrule
 & P & R & $F_{0.5}$ & P & R & $F_{0.5}$ \\ \midrule
Transformer & 37.69 & \textbf{37.67} & 37.72 & 67.27 & \textbf{42.05} & \textbf{60.06} \\
LSTM & \textbf{48.68} & 29.37 & \textbf{43.02} & \textbf{72.97} & 31.09 & 57.47 \\
CNN & 44.35 & 30.87 & 40.78 & 70.85 & 32.77 & 57.49 \\
SMT & 40.73 & 18.60 & 32.91 & 67.95 & 16.89 & 42.35 \\
 \bottomrule
\end{tabular}
 }
 \caption{Performance in precision, recall, and $F_{0.5}$ of all models on the corpora when the WER is lowest and highest.}
 \label{rpf_low_high}
\end{table}


\subsection{Is Diverse Single-Corpus Evaluation Sufficient?}

Experimental results indicate that the benchmark single-corpus evaluation is not robust; however, more diverse corpora remain undetermined. Both JFLEG and FCE can be diverse corpora because they contain examination scripts written by language learners from all over the world.
JFLEG is particularly designed to contain more diverse corpus for developing and evaluating GEC models \cite{Napoles:17:EACL}.
If a diverse single-corpus evaluation suffices, the rankings of the models will remain the same.
However, experimental results have shown that the model rankings on both JFLEG and FCE are different (Figure~\ref{f05_ranking}).
Thus, single-corpus evaluation is deemed weak regardless of its diversity.

\subsection{Advantage of Cross-Corpora Evaluation}

This study discusses the importance of evaluating GEC models from various perspectives using multiple corpora. Multi-perspective evaluation does not necessarily mean using multiple corpora. Many aspects in a corpus can be used for analysis, such as the proficiency of the writers, essay topics, and the writer 's native language. As a case study, we evaluate and analyze the models regarding the essay WER. Table \ref{rpf_low_high} shows the performance (in precision, recall, and $F_{0.5}$) of all the models when WER is the lowest (7.64 \% for ICNALE) and the highest (20.86 \% for JFLEG). Transformer and LSTM outperform all the other models in the highest and the lowest error-rated corpora, respectively. Experimental results show that LSTM and transformer may be more precision-oriented and recall-oriented, respectively. Further, precision-oriented models have an advantage over recall-oriented models when a given text contains several errors, and vice versa. This knowledge enables choosing a model based on the task that has to be completed.

\section{Conclusion}

This study explored the necessity of performing cross-corpora evaluation for GEC models, for which the performance of several GEC models was investigated against various learner corpora. Empirical evaluation results revealed that the model performance and rankings considerably vary depending on the corpus, suggesting that a single-corpus evaluation can be unreliable. Therefore, cross-corpora evaluation should be applied to GEC models. We also published our source code for the cross-corpora evaluation framework so that researchers in the community can adequately and easily evaluate their models based on multiple corpora. Our future study will further examine the robustness of several existing evaluation metrics and explore new metrics appropriate for cross-corpora and/or cross-domain evaluation.

\section*{Acknowledgments}
We are grateful to the members of the Tohoku University Natural Language Processing Laboratory as well as the anonymous reviewers for their insightful comments and suggestions.

\bibliographystyle{acl_natbib}
\bibliography{main}

\end{document}